\documentclass{article}[12pt]

\usepackage{graphicx}
\usepackage{latexsym}
\usepackage{amsmath}
\usepackage[psamsfonts]{amssymb}

\newtheorem{Theorem}{{\bf Theorem}}
\newtheorem{Lemma}{{\bf Lemma}}

\title{Equations of States in Statistical Learning \\
for a Nonparametrizable and Regular Case}

\author{Sumio Watanabe\\
PI Lab., Tokyo Institute of Technology, \\
Mailbox R2-5, 4259,
Midori-ku, Yokohama, 226-8503 Japan\\
(E-mail) swatanab(AT)pi.titech.ac.jp
}

\begin{document}

\large

\setlength{\baselineskip}{18pt}

\maketitle

\begin{abstract}
Many learning machines that have hierarchical structure or hidden variables are
now being used in information science, artificial intelligence, and bioinformatics. 
However, several learning machines used in such fields are not regular but 
singular statistical models, 
hence their generalization performance is still left unknown. To overcome these problems, 
in the previous papers, we proved new equations in statistical learning,  
by which we can estimate the Bayes generalization loss
from the Bayes training loss and the functional variance, on the condition that
the true distribution is
a singularity contained in a learning machine. 
In this paper, we prove that the same equations hold even if a 
true distribution is not contained in a parametric model. 
Also we prove that, the proposed equations in 
a regular case are asymptotically equivalent to the Takeuchi information criterion. 
Therefore, the proposed equations are always applicable without any condition 
on the unknown true distribution. 
\end{abstract}

\section{Introduction}

Nowadays, a lot of learning machines are being used in information science,
artificial intelligence, and bioinformatics. However, several learning machines
used in such fields, for example, 
three-layer neural networks, hidden Markov models, normal mixtures,
binomial mixtures, Boltzmann machines, and reduced rank regressions
have hierarchical structure or hidden variables, with the result that
the mapping from the parameter to the probability distribution is not
one-to-one. In such learning machines, it was pointed out that 
the maximum likelihood estimator is not subject to the normal distribution
\cite{Hartigan, Hagiwara, Hayasaka, Amari}, 
and that the {\it a posteriori} distribution can not be approximated by any gaussian distribution
\cite{1995,2001a,2001b,2001c}. Hence the conventional statistical 
methods for model selection, hypothesis test, and hyperparameter optimization 
are not applicable to such learning machines. 
In other words, we have not yet established the theoretical foundation
for learning machines which extract hidden structures from random samples. 

In statistical learning theory, we study the problem of learning and
generalization based on several assumptions. 
Let $q(x)$ be a true probability density function and 
$p(x|w)$ be a learning machine, which is represented by 
a probability density function of $x$ for a parameter $w$. 
In this paper, we examine the following two assumptions. \\
(1) The first is the parametrizability condition. 
A true distribution $q(x)$ is said to be {\it parametrizable}
by a learning machine $p(x|w)$, 
if there is a parameter $w_{0}$ which
satisfies $q(x)=p(x|w_{0})$. If 
otherwise, it is called {\it nonparametrizable}.\\
(2) The second is the regularity condition. 
A true distribution $q(x)$ is said to be {\it regular} for a learning machine $p(x|w)$, 
if the parameter $w_{0}$ that minimizes the log loss function
\begin{equation}\label{eq:L(w)}
L(w)=-\int q(x) \log p(x|w) dx
\end{equation}
is unique and if the Hessian matrix $\nabla^{2} L(w_{0})$ is positive definite.
If a true distribution is not regular for a learning machine, then it is said to be 
{\it singular}.

In study of layered neural networks and normal mixtures, 
both conditions are important. In fact, 
if a learning machine is redundant compared to a true distribution, then
the true distribution is parametrizable and singular. 
Or if a learning machine is too simple to approximate a 
true distribution, then the true distribution is nonparametrizable and 
regular. In practical applications, we need a method 
to determine the optimal learning machine, 
therefore, a general formula is desirable by which
the generalization loss can be estimated from the training loss
without regard to such conditions. 

In the previous papers \cite{2007,NC2007,WCCI2008,MSJ2009,Cambridge}, 
we studied a case when a true distribution is parametrizable and singular, 
and proved new formulas which enable us 
to estimate the generalization loss from the training loss and the functional variance. 
Since the new formulas hold for an arbitrary 
set of a true distribution,  a learning machine, and
an {\it a priori} distribution, they are called 
{\it equations of states in statistical estimation}. 
However, it has not been clarified whether they hold or not 
in a nonparametrizable case. 

In this paper, we study the case when a true distribution is nonparametrizable and regular, and 
prove that the same equations of states also hold. Moreover, we show that, in 
a nonparametrizable and regular case, the equations of states are asymptotically 
equivalent to the Takeuchi information criterion (TIC) 
for the maximum likelihood method. Here TIC was 
derived for the model selection criterion in the case when the true distribution 
is not contained in a statistical model \cite{Takeuchi}. The network information criterion
\cite{Murata}
was devised by generalizing it to an arbitrary loss function in the regular case.

If a true distribution is singular for a learning machine, 
TIC is ill-defined, whereas the equations of states
are well-defined and equal to the average generalization losses. 
Therefore, equations of states can be understood as the generalized version of
TIC from the maximum likelihood method in a regular case
to Bayes method for regular and singular cases. 

This paper consists of six sections. In Section 2, we summarized the
framework of Bayes learning and the results of previous papers. In Section 3, 
we show the main results of this paper. In Section 4, some lemmas are prepared 
which are used in the proofs of the main results. The proofs of lemmas are
given in the Appendix. In Section 5, we prove the main theorems. 
In Section 5 and 6, we discuss and conclude this paper. 

\section{Background}

In this section, we summarize the background of the paper. 

\subsection{Bayes learning}

Firstly we introduce the framework of Bayes and Gibbs estimations,
which is well known in statistics and learning theory. 

Let $N$ be a natural number and ${\bf R}^{N}$ be the $N$-dimensional Euclidean space.
Assume that an information source is given by 
a probability density function $q(x)$ on ${\bf R}^{N}$ 
and that random samples $X_{1},X_{2},...,X_{n}$
are independently subject to the probability distribution $q(x)dx$. 
Sometimes $X_{1},X_{2},..,X_{n}$ are said to be training samples and 
the information source $q(x)$ is called a true probability density function. 
In this paper we use notations for a given function $g(x)$, 
\begin{eqnarray*}
E_{X}[g(X)]&=&\int g(x)q(x)dx,\\
E_{j}^{(n)}[g(X_{j})]&=&\frac{1}{n}\sum_{j=1}^{n}g(X_{j}).
\end{eqnarray*}
Note that the expectation $E_{X}[g(X)]$ is given by the integration by 
the true distribution, but that 
the empirical expectation $E_{j}^{(n)}[g(X_{j})]$ can be calculated using
random samples. 

We study a learning machine $p(x|w)$ of $x\in {\bf R}^{N}$
for a given parameter $w\in {\bf R}^{d}$. 
Let $\varphi(w)$ be an {\it a priori} probability density function on ${\bf R}^{d}$. 
The expectation operator $E_{w}[\;\;]$ by 
the {\it a posteriori} probability distribution with the inverse temperature 
$\beta>0$ for a given function $g(w)$ is defined by 
\[
E_{w}[g(w)]=\frac{1}{Z(\beta)}\int g(w) \varphi(w)\prod_{i=1}^{n}p(X_{i}|w)^{\beta}\;dw,
\]
where $Z(\beta)$ is the normalizing constant. 
The Bayes generalization loss $B_{g}$, the Bayes training loss $B_{t}$, 
the Gibbs generalization loss $G_{g}$, and the Gibbs training loss $G_{t}$
are respectively defined by
\begin{eqnarray*}
B_{g}&=& - E_{X}[\;\log E_{w}[p(X|w)]\;],\\
B_{t}&=& - E_{j}^{(n)}[\;\log E_{w}[p(X_{j}|w)]\;], \\
G_{g}&=& - E_{X}[\;E_{w}[\log p(X|w)]\;],\\
G_{t}&=& - E_{j}^{(n)}[\;E_{w}[\log p(X_{j}|w)]\;].
\end{eqnarray*}
The functional variance $V$ is defined by 
\[
V =  n \times E_{j}^{(n)}\{
E_{w}[\;(\log p(X_{j}|w))^{2}\;]
-E_{w}[\log p(X_{j}|w)]^{2}\}.
\]
The concept of the functional variance was
firstly proposed in the papers \cite{2007,NC2007,WCCI2008,MSJ2009}.
In this paper, we show that the functional variance plays an important role in 
learning theory. 
Remark that $B_{g}$, $B_{t}$, $G_{g}$, $G_{t}$, and $V$ are random variables because 
$E_{w}[\;\;]$ depends on random samples. 
Let $E[\;\;]$ denote the expectation value overall sets of 
training samples. Then $E[B_{g}]$ and $E[B_{t}]$ are respectively called
the average Bayes generalization and training error, and $E[G_{g}]$ and $E[G_{t}]$ 
the average Gibbs ones. 

In theoretical analysis, we assume some conditions on a true distribution and
a learning machine. If there exists a parameter $w_{0}$ such that $q(x)=p(x|w_{0})$,
then the true distribution is said to be 
parametrizable. If otherwise, 
nonparametrizable. In both cases, we define $w_{0}$ as the parameter that minimizes
the log loss function $L(w)$ in eq.(\ref{eq:L(w)}). 
Note that $w_{0}$ is equal to the parameter that minimizes the Kullback-Leibler
distance from the true distribution to the parametric model. 
If $w_{0}$ is unique and if the Hessian matrix 
\[
\frac{\partial^{2}}{\partial w_{j}\partial w_{k}} L(w_{0})
\]
is positive definite, then the true distribution is said to be regular for
a learning machine. 
\vskip3mm\noindent{\bf Remark.}
Several learning machines such as a layered neural network or a normal mixture
have natural nonidentifiability by the symmetry of a parameter. For example,
in a normal mixture, 
\[
p(x|a,b,c)=\frac{a}{\sqrt{2\pi}}\;e^{-|x-b|^{2}/2}
+ \frac{1-a}{\sqrt{2\pi}}\;e^{-|x-c|^{2}/2},
\]
two probability distributions $p(x|a,b,c)$ and $p(1-a,c,b)$ give the same 
probability distribution, hence the parameter $w_{0}$ that minimizes $L(w)$ is
not unique for any true distribution. 
In a parametrizable and singular case, such nonidentifiability strongly 
affects learning \cite{1995,2001a}. However, in a nonparametrizable and regular case, 
the {\it a posteriori} distribution in the neighborhood of 
each optimal parameter has the same form, resulting that we can assume $w_{0}$ 
is unique without loss of generality. 

\subsection{Notations}

Secondly, we explain some notations. 

For given scalar functions $f(w)$ and $g(w)$, 
the vector $\nabla f(w)$ and two matrices
$\nabla f(w)\nabla g(w)$ and $\nabla^{2}f(w)$ are respectively defined by 
\begin{eqnarray*}
(\nabla f(w))_{j}&=&\frac{\partial f(w)}{\partial w_{j}},\\
(\nabla f(w)\nabla g(w))_{jk}&=&
\frac{\partial f(w)}{\partial w_{j}}
\frac{\partial g(w)}{\partial w_{k}},\\
(\nabla^{2} f(w))_{jk}&=&
\frac{\partial^{2} f(w)}{\partial w_{j}\partial w_{k}}.
\end{eqnarray*}

Let $n$ be the number of training samples. 
For a given constant $\alpha$, we use the following notations.\\
(1) $Y_{n}=O_{p}(n^{\alpha})$ shows that a random variable $Y_{n}$
satisfies $|Y_{n}|\leq C n^{\alpha}$ with some random variable $C\geq 0$. \\
(2) $Y_{n}=o_{p}(n^{\alpha})$ shows that a random variable $Y_{n}$ 
satisfies the convergence in probability 
$|Y_{n}|/n^{\alpha}\rightarrow 0$. \\
(3) $y_{n}=O(n^{\alpha})$ shows that a sequence $y_{n}$
satisfies $|y_{n}|\leq C n^{\alpha}$ with some constant $C\geq 0$. \\
(4) $y_{n}=o(n^{\alpha})$ shows that a sequence $y_{n}$ 
satisfies the convergence $|y_{n}|/n^{\alpha}\rightarrow 0$. 
\vskip3mm
\noindent{\bf Remark.}
For a sequence of random variables, 
it needs mathematically technical procedure 
to prove convergence in probability or convergence in law. 
If we adopt the completely mathematical
procedure in the proof, a lot of readers in information science 
may not find the essential points in the theorems. 
For example, see \cite{2007,MSJ2009,Cambridge}. Therefore, in this paper, we adopt 
the natural and appropriate level of mathematical rigorousness, 
from the viewpoint of mathematical sciences. 
The notations $O_{p}$ and $o_{p}$ are very useful and understandable for such a purpose.

\subsection{Parametrizable and singular case}

Thirdly, we introduce the results of the 
previous researches \cite{2007,NC2007,WCCI2008,MSJ2009}. 
We do not prove these results in this paper. 

Assume that a true distribution is parametrizable. Even if the true distribution
is singular for a learning machine, 
\begin{eqnarray}
E[B_{g}]&=&S_{0}+\frac{\lambda_{0}-\nu_{0}}{n\beta}+\frac{\nu_{0}}{n}+o(\frac{1}{n}),\label{eq:11a}\\
E[B_{t}]&=&S_{0}+\frac{\lambda_{0}-\nu_{0}}{n\beta}-\frac{\nu_{0}}{n}+o(\frac{1}{n}),\label{eq:22a}\\
E[G_{g}]&=&S_{0}+\frac{\lambda_{0}}{n\beta}+\frac{\nu_{0}}{n}+o(\frac{1}{n}),\label{eq:11b}\\
E[G_{t}]&=&S_{0}+\frac{\lambda_{0}}{n\beta}-\frac{\nu_{0}}{n}+o(\frac{1}{n}),\label{eq:22b}\\
E[V]&=&\frac{2\nu_{0}}{\beta}+o(1), \label{eq:33a}
\end{eqnarray}
where $S_{0}$ is the entropy of the true probability density function $q(x)$, 
\[
S_{0}=-\int q(x)\log q(x)dx.
\]
The constants $\lambda_{0}$ and $\nu_{0}$ are respectively
the generalized log canonical threshold and the singular fluctuation,
which are birational invariants. The concrete values of them can be derived by
using algebraic geometrical transformation called resolution of singularities. 
By elliminating $\lambda_{0}$ and $\nu_{0}$ from eq.(\ref{eq:11a})-eq.(\ref{eq:33a}), 
\begin{eqnarray}
E[B_{g}]&=&E[B_{t}]+(\beta/n) E[V]+o(\frac{1}{n})\label{eq:maina},\\
E[G_{g}]&=&E[G_{t}]+(\beta/n) E[V]+o(\frac{1}{n})\label{eq:mainb},
\end{eqnarray}
hold, which are called {\it equations of states in learning},
because these relations hold for an arbitrary set of a true distribution,
a learning machine, and an {\it a priori} distribution.
By this relation, we can estimate the generalization loss 
using the training loss and the functional variance. 
However, it has been left unknown whether
the equations of states, eq.(\ref{eq:maina}) and eq.(\ref{eq:mainb}), 
hold or not in nonparametrizable cases. 

\section{Main Results}

In this section, we describe the main results of this paper. 
The proofs of theorems are given in Section 5. 

\subsection{Equations of states} 

In this paper, study the case when a true distribution is 
nonparametrizable and regular.
Three constants $S$, $\lambda$, and $\nu$ are respectively 
defined by the following equations. 
Let $w_{0}$ be the unique parameter that minimizes $L(w)$. 
Three constants are defined by 
\begin{eqnarray}
S &=& L(w_{0}),\\
\lambda&=& \frac{d}{2}, \\
\nu&=&\frac{1}{2}\mbox{tr}(IJ^{-1})\label{eq:nu},
\end{eqnarray}
where $d$ is the dimension of the parameter, and 
$I$ and $J$ are $d\times d$ matrices defined by 
\begin{eqnarray}
I&=&\int\nabla\log p(x|w_{0})
\nabla\log p(x|w_{0}) q(x)dx,\label{eq:Iij}\\
J&=&- \int\nabla^{2}\log p(x|w_{0})
 q(x)dx.\label{eq:Jij}
\end{eqnarray}

\begin{Theorem} \label{Theorem:111}
Assume that a true distribution $q(x)$ is nonparametrizable and regular for 
a learning machine $p(x|w)$. Then
\begin{eqnarray}
E[B_{g}]&=&S+\frac{\lambda-\nu}{n\beta}+\frac{\nu}{n}+o(\frac{1}{n}),\label{eq:111a}\\
E[B_{t}]&=&S+\frac{\lambda-\nu}{n\beta}-\frac{\nu}{n}+o(\frac{1}{n}),\label{eq:222a}\\
E[G_{g}]&=&S+\frac{\lambda}{n\beta}+\frac{\nu}{n}+o(\frac{1}{n}),\label{eq:111b}\\
E[G_{t}]&=&S+\frac{\lambda}{n\beta}-\frac{\nu}{n}+o(\frac{1}{n}),\label{eq:222b}\\
E[V]&=&\frac{2\nu}{\beta}+o(1).\label{eq:333a}
\end{eqnarray}
Therefore, equations of states hold, 
\begin{eqnarray}
E[B_{g}]&=&E[B_{t}]+(\beta/n) E[V]+o(\frac{1}{n})\label{eq:maina2},\\
E[G_{g}]&=&E[G_{t}]+(\beta/n) E[V]+o(\frac{1}{n})\label{eq:mainb2}.
\end{eqnarray}
\end{Theorem}
Proof of this theorem is given in Section 5. 
Note that constants are different between the parametrizable and nonparametrizable cases, 
that is to say, $S\neq S_{0}$, $\lambda\neq \lambda_{0}$, 
and $\nu\neq \nu_{0}$. However, the same equations of states still hold. In fact, 
eq.(\ref{eq:maina2}) and eq.(\ref{eq:mainb2})
are completely equal to as eq.(\ref{eq:maina}) and eq.(\ref{eq:mainb}),
respectively. 

By combining the results of the previous papers with the new result in Theorem 1, 
it is ensured that the equations of states
are applicable to arbitrary set of a true distribution, 
a learning machine, and an {\it a priori} distribution,
without regard to the condition on the unknown true distribution. 
\vskip3mm\noindent
{\bf Remark}. 
If a true distribution is parametrizable and regular, 
then  $I=J$, hence $\lambda=\nu=d/2$. If otherwise,
$I\neq J$ in general.  
Note that $J$ is positive definite by the assumption, 
but that $I$ may not be positive definite in general. 

\subsection{Comparison TIC with equations of states}

If the maximum likelihood method is employed, or equivalently if $\beta=\infty$, 
then $B_{g}$ and $B_{t}$ are 
respectively equal to the generalization and training losses of the maximum likelihood 
method. It was proved in \cite{Takeuchi} that  
\begin{equation}\label{eq:TIC-intro}
E[B_{g}]=E[B_{t}]+\frac{TIC}{n}+o(\frac{1}{n})\;\;\;(\beta=\infty),
\end{equation}
where 
\[
TIC=\mbox{tr}(I(w_{0})J(w_{0})^{-1}).
\]
On the other hand,  the equations of states, eq.(\ref{eq:maina2}) in Theorem 1
show that,
\begin{equation}\label{eq:TIC-intro-2}
E[B_{g}]=E[B_{t}]+\frac{E[\beta V]}{n}+o(\frac{1}{n}).
\;\;\;(0<\beta< \infty),
\end{equation}
Therefore, in this subsection, let us compare $\beta V$ with 
$TIC$ in the nonparametrizable and regular case. 

Let $L_{n}(w)$ be the empirical log loss function
\[
L_{n}(w)=-E_{j}^{(n)}[\log p(X_{j}|w)]-\frac{1}{n\beta}\log\varphi(w).
\]
Three matrices are defined by
\begin{eqnarray}
I_{n}(w)&=&E_{j}^{(n)}[\nabla\log p(X_{j}|w)
\nabla\log p(X_{j}|w)],\label{eq:Inij}\\
J_{n}(w)&=&-E_{j}^{(n)}[\nabla^{2}\log p(X_{j}|w)],\label{eq:Jnij}\\
K_{n}(w)&=&\nabla^{2} L_{n}(w).\label{eq:Knij}
\end{eqnarray}
In practical applications, instead of $TIC$, 
the empirical TIC is employed, 
\[
TIC_{n}=\mbox{tr}(I_{n}(w_{MLE})J_{n}(w_{MLE})^{-1}),
\]
where $w_{MLE}$ is the maximum likelihood estimator. 
Then by using the convergence in probability 
$w_{MLE}\rightarrow w_{0}$, 
\[
E[TIC_{n}]=TIC+o(1).
\]
On the other hand, we have shown in Theorem \ref{Theorem:111}, 
\[
E[\beta V]=TIC+o(1).
\]
Hence let us compare $\beta V$ with $TIC_{n}$ as random variables. 
\begin{Theorem}\label{Theorem:222}
Assume that $q(x)$ is nonparametrizable and regular for
a learning machine $p(x|w)$. Then 
\begin{eqnarray*}
TIC_{n}&=& TIC+O_{p}(\frac{1}{\sqrt{n}}), \\
\beta V &=& TIC +O_{p}(\frac{1}{\sqrt{n}}),\\
\beta V&=& TIC_{n}+O_{p}(\frac{1}{n}).
\end{eqnarray*}
\end{Theorem}
Proof of this theorem is given in Section 5. 
Theorem 2 shows that the difference between $\beta V$ and
$TIC_{n}$ is in the smaller order than the variance of them. 
Therefore, if a true distribution is nonparametrizable and regular
for a learning machine, then the equations of states are asymptotically
equivalent to the empirical TIC. If a true distribution is singular or 
if the number of training samples are not so large, then 
the empirical TIC and the equations of states are not
equivalent, in general. Hence the equations of states are
applicable more widely than TIC. 
Experimental analysis for the equations of states was reported in 
\cite{2007,NC2007,WCCI2008}. The main purpose of this paper is
to prove Theorems 1 and 2. Its application to practical problems
is a topic for the future study.

\section{Preparation of Proof}

In this section, we summarize the basic properties 
which are used in the proofs of main theorems.

\subsection{Maximum {\it a posteriori} estimator}

Firstly, we study the asymptotic behavior of the maximum 
{\it a posteriori} estimator. By the definition, for each $w$, 
\[
K_{n}(w)=J_{n}(w)+O(\frac{1}{n}).
\]
By the central limit theorem, for each $w$,  
\begin{eqnarray}
I_{n}(w)& = & I(w)+O_{p}(\frac{1}{\sqrt{n}}),\label{eq:In}\\
J_{n}(w)& = & J(w)+O_{p}(\frac{1}{\sqrt{n}}),\label{eq:Jn}\\
K_{n}(w)& = & J(w)+O_{p}(\frac{1}{\sqrt{n}}).\label{eq:Kn}
\end{eqnarray}
The parameter that minimizes $L_{n}(w)$ is denoted by $\hat{w}$,
which is called the maximum {\it a posteriori} estimator (MAP). 
If $\beta=1$, then it is equal to the conventional maximum {\it a posteriori}
estimator (MAP). If $\beta=\infty$, or equivalently $1/\beta=0$, 
then it is the maximum likelihood estimator (MLE),
which is denoted by $w_{MLE}$.  

Let us summarize the basic properties of the maximum 
{\it a posteriori} estimator.
Becaue $w_{0}$ and $\hat{w}$ minimizes $L(w)$ and $L_{n}(w)$ respectively, 
\begin{eqnarray}
\nabla L(w_{0})&=&0,\label{eq:nabla-0-1}\\
\nabla L_{n}(\hat{w})&=&0.\label{eq:nabla-0-2}
\end{eqnarray}
By the assumption, $w_{0}$ is unique and the matrix $J$ is positive definite, 
the consistency of $\hat{w}$ holds under the natural condition, in other words,
the convergences in probability
$\hat{w}\rightarrow w_{0}$ ($n\rightarrow\infty$) hold for $0<\beta\leq \infty$. 
In this paper, we assume such consistency condition. 

From eq.(\ref{eq:nabla-0-2}), there exists $w_{\beta}^{*}$ 
which satisfies 
\begin{equation}\label{eq:nablaln}
\nabla L_{n}(w_{0})+\nabla^{2}L_{n}(w_{\beta}^{*})(\hat{w}-w_{0})=0
\end{equation}
and 
\[
|w_{\beta}^{*}-w_{0}|\leq |\hat{w}-w_{0}|,
\]
where $|\cdot|$ denotes the norm of ${\bf R}^{d}$. 
By using the definition $K_{n}(w_{\beta}^{*})=\nabla^{2} L_{n}(w_{\beta}^{*})$, 
\begin{equation}\label{eq:w*-w0}
\hat{w}-w_{0}=-K_{n}(w_{\beta}^{*})^{-1}\nabla L_{n}(w_{0}). 
\end{equation}
By using the law of large numbers and the central limit theorem, 
$K_{n}(w_{\beta}^{*})$ converges to
$J$ in probability and $\sqrt{n}\;\nabla L_{n}(w_{0})$ 
converges in law to the normal distribution with average 0
and covariance matrix $I$. Therefore
\[
\sqrt{n}\;(\hat{w}-w_{0})
\]
converges in law to the normal distribution with average 0
and covariance matrix $J^{-1}IJ^{-1}$,
resulting that 
\begin{equation}
E[(\hat{w}-w_{0})(\hat{w}-w_{0})^{T}]=\frac{J^{-1}IJ^{-1}}{n}
+o(\frac{1}{n}),\label{eq:JIJ}
\end{equation}
for $0<\beta\leq \infty$, where $(\;\;)^{T}$ denotes the transposed vector. 
In other words, 
\begin{equation}
\hat{w}=w_{0}+O_{p}(\frac{1}{\sqrt{n}}).
 \label{eq:MAP-TRUE}
\end{equation}
Hence,
\[
K_{n}(w_{\beta}^{*})=J(w_{0})+O_{p}(\frac{1}{\sqrt{n}}).
\]
By using eq.(\ref{eq:w*-w0}), 
\[
\hat{w}-w_{MLE}=
\Bigl(K_{n}(w_{\infty}^{*})^{-1}-K_{n}(w_{\beta}^{*})^{-1}
\Bigr)\nabla L_{n}(w_{0}).
\]
Since $\nabla L_{n}(w_{0})=O_{p}(1/\sqrt{n})$ and $J(w_{0})$ is positive definite, 
we have
\begin{equation}
w_{MLE}=\hat{w}+O_{p}(\frac{1}{n}).\label{eq:MAP-MLE}
\end{equation}

\subsection{Expectations by {\it a posteriori} distribution}

Secondly, the behavior of the {\it a posteriori} distribution 
is described as follows. 

For a given function $g(w)$, 
the average by the {\it a posteriori} distribution is defined by 
\[
E_{w}[g(w)]=\frac{ \int g(w) 
\exp(-n\beta L_{n}(w))dw} 
{ \int \exp(-n\beta L_{n}(w))dw}.
\]
Then we can prove the following relations. 
\begin{Lemma}\label{Lemma:111}
\begin{eqnarray}
E_{w}[(w-\hat{w})] & =& O_{p}(\frac{1}{n}),\label{eq:w-w0} \\
E_{w}[(w-\hat{w})(w-\hat{w})^{T}] & = & 
\frac{K_{n}(\hat{w})^{-1}}{n\beta}
+O_{p}(\frac{1}{n^{2}}),\label{eq:w-w0-2} \\
E_{w}[(w-\hat{w})_{i}(w-\hat{w})_{j}
(w-\hat{w})_{k}] & =& O_{p}(\frac{1}{n^{2}}),\label{eq:w-w0-3} \\
E_{w}[|w-\hat{w}|^{m}] & =& O_{p}(\frac{1}{n^{m/2}})\;\;\;(m\geq 1).\label{eq:w-w0-4} 
\end{eqnarray}
Moreover, 
\begin{eqnarray}
EE_{w}[(w-w_{0})(w-w_{0})^{T}] & = & 
\frac{J^{-1}IJ^{-1}}{n}+\frac{J^{-1}}{n\beta}
+o(\frac{1}{n}). \label{eq:e(w-0)(w-0)}\\
EE_{w}[|w-w_{0}|^{3}] & = & o(\frac{1}{n}). \label{eq:e(w-0)3}
\end{eqnarray}
\end{Lemma}
For the proof of this lemma, see Appendix. 

Let us introduce a log density ratio function $f(x,w)$ by
\[
f(x,w)=\log\frac{p(x|w_{0})}{p(x|w)}.
\]
Then $f(x,w_{0})\equiv 0$ and 
\begin{eqnarray*}
\nabla f(x,w)&=&-\nabla \log p(x|w),\\
\nabla^{2} f(x,w)&=& - \nabla^{2}\log p(x|w).
\end{eqnarray*}
In the proof of Theorems 1, we need the following six expectation values, 
\begin{eqnarray*}
D_{1}&=& E E_{X}[E_{w}[f(X,w)]],\\
D_{2}&=& (1/2)EE_{X}[E_{w}[\;f(X,w)^{2}\;]],\\
D_{3}&=& (1/2)EE_{X}[\;E_{w}[f(X,w)]^{2}\;],\\
D_{4}&=& E E_{j}^{(n)}[E_{w}[f(X_{j},w)]],\\
D_{5}&=& (1/2)E E_{j}^{(n)}[E_{w}[\;f(X_{j},w)^{2}\;]],\\
D_{6}&=& (1/2)E E_{j}^{(n)}[\;E_{w}[f(X_{j},w)]^{2}\;].
\end{eqnarray*}
The constant $\mu$ is defined by 
\begin{equation}\label{eq:mu}
\mu=\frac{1}{2}\mbox{tr}(IJ^{-1}IJ^{-1}).
\end{equation}
Then we can prove the following relations. 
\begin{Lemma}\label{Lemma:222} 
Let $\nu$ and $\mu$ be constants which are respectively defined by
eq.(\ref{eq:nu}) and eq.(\ref{eq:mu}). Then 
\begin{eqnarray*}
D_{1}&=&\frac{d}{2n\beta}+\frac{\nu}{n}+o(\frac{1}{n}), \\
D_{2}&=&\frac{\nu}{n\beta}+\frac{\mu}{n}+o(\frac{1}{n}), \\
D_{3}&=&\frac{\mu}{n}+o(\frac{1}{n}), \\
D_{4}&=&\frac{d}{2n\beta}-\frac{\nu}{n}+o(\frac{1}{n}), \\
D_{5}&=&\frac{\nu}{n\beta}+\frac{\mu}{n}+o(\frac{1}{n}), \\
D_{6}&=&\frac{\mu}{n}+o(\frac{1}{n}). 
\end{eqnarray*}
\end{Lemma}
For the proof of this lemma, see Appendix. 

\section{Proofs}

In this section, we prove theorems. 

\subsection{Proof of Theorem \ref{Theorem:111}}

Firstly, by using the definitions
\begin{eqnarray*}
S & = & L(w_{0})=-E_{X}[\log p(X|w_{0})],\\
p(x|w)& = & p(x|w_{0})\exp(-f(x,w)),
\end{eqnarray*}
the Bayes generalization loss is given by 
\begin{eqnarray*}
E[B_{g}]&=&- EE_{X}\log E_{w}[p(X|w)] \\
&=& S-EE_{X}[\log E_{w}[\exp(-f(X,w))]]\\
&=&S-EE_{X}[\log E_{w}(1-f(X,w)+\frac{f(X,w)^{2}}{2})]+o(\frac{1}{n}) \\
&=&S+EE_{X}E_{w}[f(X,w)] -\frac{1}{2}EE_{X}E_{w}[f(X,w)^{2}] \\
& & + \frac{1}{2}EE_{X}[\;E_{w}[f(X,w)]^{2}\;]+o(\frac{1}{n}) \\
&=&S+D_{1}-D_{2}+D_{3} +o(\frac{1}{n}) \\
&=& S+\frac{d}{2n\beta}-\frac{\nu}{n\beta}+
\frac{\nu}{n}+o(\frac{1}{n}).
\end{eqnarray*}
Secondly, the Bayes training loss is 
\begin{eqnarray*}
E[B_{t}]&=& - EE_{j}^{(n)}\log E_{w}[p(X_{j}|w)] \\
&=&S-EE_{j}^{(n)}[\log E_{w}[\exp(-f(X_{j},w))] ]\\
&=& S-EE_{j}^{(n)}[\log E_{w}(1-f(X_{j},w)+\frac{f(X_{j},w)^{2}}{2})]+o(\frac{1}{n}) \\
&=& S+EE_{j}^{(n)}E_{w}[f(X_{j},w)] -\frac{1}{2}EE_{j}^{(n)}E_{w}[f(X_{j},w)^{2}] \\
& &+ \frac{1}{2}EE_{j}^{(n)}[\;E_{w}[f(X_{j},w)]^{2}\;]+o(\frac{1}{n}) \\
&=& S+D_{4}-D_{5}+D_{6} +o(\frac{1}{n}) \\
&=& S+\frac{d}{2n\beta}-\frac{\nu}{n\beta}-
\frac{\nu}{n}+o(\frac{1}{n}).
\end{eqnarray*}
Thirdly, the Gibbs generalization loss is 
\begin{eqnarray*}
E[G_{g}]&=& - EE_{X}E_{w}[\log p(X|w)] \\
&=& S+EE_{X}E_{w}[f(X,w)] \\
&=& S+D_{1} \\
&=& S+\frac{d}{2n\beta}+\frac{\nu}{n}+o(\frac{1}{n}).
\end{eqnarray*}
Forthly, the Gibbs training loss is 
\begin{eqnarray*}
E[G_{t}]&=& - EE_{X}^{(n)}E_{w}[\log p(X|w)] \\
&=& S+EE_{X}^{(n)}E_{w}[f(X,w)] \\
&=& S+D_{4}\\
&=& S+\frac{d}{2n\beta}-\frac{\nu}{n}+o(\frac{1}{n}).
\end{eqnarray*}
Lastly, the functional variance is given by 
\begin{eqnarray*}
E[V]&=& 2n(D_{5}-D_{6}) \\
&=& 2n(D_{2}-D_{3})+o(1) \\
&=& \frac{2\nu}{\beta}+o(1). 
\end{eqnarray*}
Therefore, we obtained Theorem 1. 

\subsection{Proof of Theorem \ref{Theorem:222}}

Let $V_{w}[f(X,w)]$ be the variance of $f(X,w)$ in the 
{\it a posteriori} distribution, 
\[
V_{w}[f(X,w)] \equiv E_{w}[f(X,w)^{2}]-E_{w}[f(X,w)]^{2}.
\]
Then 
\[
V_{w}[f(X,w)]=V_{w}[f(X,w)-f(X,\hat{w})]
\]
holds because $f(X,\hat{w})$ is a constant function of $w$. 
By the Taylor expansion at $w=\hat{w}$, 
\begin{eqnarray*}
&& f(X,w) - f(X,\hat{w})= \nabla f(X,\hat{w})\cdot (w-\hat{w}) \\
&&+\frac{1}{2}(w-\hat{w})\cdot \nabla^{2} f(X,\hat{w})(w-\hat{w}) 
+O(|w-\hat{w}|^{3}).
\end{eqnarray*}
Using this expansion, and 
eq.(\ref{eq:w-w0}), eq.(\ref{eq:w-w0-2}), eq.(\ref{eq:w-w0-3}), 
and eq.(\ref{eq:w-w0-4}), 
\[
V_{w}[f(X,w)]=V_{w}[(\nabla f(X,\hat{w}))\cdot(w-\hat{w})]
+O_{p}(\frac{1}{n^{2}}). 
\]
Hence
\begin{eqnarray*}
\beta V &\equiv & n\beta E_{j}^{(n)}[V_{w}[f(X_{j},w)]]\\
&=&n\beta E_{j}^{(n)}\{ 
E_{w}[(\nabla f(X_{j},\hat{w})\cdot(w-\hat{w}))^{2}]\\
&&-E_{w}[\nabla f(X_{j},\hat{w})\cdot(w-\hat{w})]^{2} \} +O_{p}(\frac{1}{n}).
\end{eqnarray*}
The second term is $O_{p}(1/n)$ by eq.(\ref{eq:w-w0}). 
Therefore, by applying eq.(\ref{eq:w-w0-2}) to 
the first term, 
\begin{eqnarray*}
\beta V&=&n\beta\;\mbox{tr}\Bigl(
E_{j}^{(n)}[(\nabla f(X_{j},\hat{w}))(\nabla f(X_{j},\hat{w}))^{T}] \\
&&\times 
E_{w}[(w-\hat{w})(w-\hat{w})^{T}]\Bigr)  +O_{p}(\frac{1}{n}) \\
&=& \mbox{tr}(I_{n}(\hat{w})K_{n}^{-1}(\hat{w}))
+O_{p}(\frac{1}{n}). 
\end{eqnarray*}
Therefore, by using eq.(\ref{eq:MAP-MLE}), 
proof of Theorem 2 is completed.

\section{Discussion}

Let us discuss the results of this paper
from the three different points of view. 

Firstly, we discuss the method how to numerically 
calculate the equations of states. 
The widely applicable information criterion (WAIC) \cite{2007,Cambridge} is 
defined by
\begin{eqnarray*}
\mbox{WAIC} & = & -\sum_{i=1}^{n}\log E_{w}[p(X_{i}|w)]\\
&& +\beta 
\sum_{i=1}^{n}
\Bigl\{
E_{w}[(\log p(X_{i}|w))^{2}]
-E_{w}[\log p(X_{i}|w)]^{2}
\Bigr\}.
\end{eqnarray*}
Then by Theorem 1, 
\[
E[WAIC]=E[nB_{g}]+o(1)
\]
holds. Hence by minimization of WAIC, we can optimize 
the model and the hyperparameter for the minimum Bayes generalization loss. 
In Bayes estimation, a set of parameters $\{w_{k}\}$ is
prepared so that it approximates the {\it a posteriori}
distribution. Sometimes it is done 
by the Markov chain Monte Carlo method, and we can approximate 
the average by the {\it a posteriori} distribution by 
\[
E_{w}[f(w)]\cong\frac{1}{K}\sum_{k=1}^{K}f(w_{k}).
\]
Therefore the WAIC can numerically calculate by such a set $\{w_{k}\}$.

Secondly, we study the fluctuation of the Bayes generalization error. 
In Theorem 1, we proved that, as the number of training samples tends to infinity,
two expectation values converge to the same value, 
\begin{eqnarray*}
E[n(B_{g}-B_{t})] & \rightarrow & \mbox{tr}(IJ^{-1}),\\
E[\beta V] & \rightarrow & \mbox{tr}(IJ^{-1}).
\end{eqnarray*}
Moreover, in Theorem 2, we proved the convergence in 
probability, 
\[
\beta V \rightarrow \mbox{tr}(IJ^{-1}).
\]
On the other hand, by the same way as Theorem 1, we can prove
\[
n(B_{g}-B_{t})=n\times \mbox{tr}(I(\hat{w}-w_{0})(\hat{w}-w_{0})^{T})+o_{p}(1).
\]
Since $\sqrt{n}(\hat{w}-w_{0})$ converges in law to the 
gaussian random variable whose
average is zero and variance is $J^{-1}IJ^{-1}$, the random variable 
$n(B_{g}-B_{t})$ converges to not a constant in probability 
but to a random variable in law.  In other words, 
the relation between expectation values 
\begin{equation}\label{eq:average-bgt}
E[B_{g}]=E[B_{t}]+\frac{\beta E[V]}{n}+o(\frac{1}{n})
\end{equation}
holds, whereas they are not equal to each other as random variables, 
\begin{equation}\label{eq:random-bgt}
B_{g}\neq B_{t}+\frac{\beta V}{n}+o_{p}(\frac{1}{n}).
\end{equation}
Note that, even if the true distribution 
is paramertrizable and regular, the generalization and training losses have 
same properties, therefore both AIC and TIC have same properties as 
eq.(\ref{eq:average-bgt}) and eq.(\ref{eq:random-bgt}). 

Lastly, let us compare the generalization loss by the Bayes estimation 
with that by the maximum likelihood estimation.
In a regular and parametrizable case, they
are equal to each other asymptotically. 
In a parametrizable and singular case, the Bayes generalization error
is smaller than that of the maximum likelihood method. 
Let us compare them in a nonparametrizable and regular case. 
\[
E[B_{g}]=S+\frac{\mbox{tr}(IJ^{-1})}{2n}+
\frac{d-\mbox{tr}(IJ^{-1})}{2n\beta}+
o(\frac{1}{n}).
\]
When $\beta=\infty$, this is the 
generalization error of the maximum likelihood method. 
If $d>\mbox{tr}(IJ^{-1})$, then 
$E[B_{g}]$ is a decreasing function of $1/\beta$. 
Or if $d<\mbox{tr}(IJ^{-1})$, then 
$E[B_{g}]$ is an increasing function of $1/\beta$. 
If $I<J$, then $\mbox{tr}(IJ^{-1})<d$.
By the definition of $I$ and $J$, 
\[
I=\int
\nabla p(x|w_{0})
\nabla p(x|w_{0})\frac{q(x)}{p(x|w_{0})^{2}}dx
\]
and
\[
J=I-Q,
\]
where
\[
Q=\int (\nabla^{2} p(x|w_{0}))\;
\frac{q(x)}{p(x|w_{0})}dx.
\]
If $Q<0$, then $\mbox{tr}(IJ^{-1})<d$, resulting that
the generalization loss of Bayes estimation is smaller than that by
the maximum likelihood method. 
\vskip3mm\noindent
{\bf Example.} For $w\in {\bf R}$, 
\[
p(x|w)=\frac{1}{\sqrt{2\pi}}
\exp(-\frac{(x-w)^{2}}{2}),
\]
Then 
\[
L(w)=\frac{1}{2}\int(x-w)^{2}q(x)dx+\frac{1}{2}\log(2\pi).
\]
Hence $w_{0}=E_{X}[X]$ and 
\[
L(w_{0})=V(X)+\frac{1}{2}\log(2\pi),
\]
where $V(X)=E_{X}[X^{2}]-E_{X}[X]^{2}$. 
The value $Q$ is
\[
Q=V(X)-1.
\]
If $V(X)>1$, then the generalization error is 
a decreasing function of  $1/\beta$, in other words,
the Bayes estimation makes the generalization loss is 
smaller than that by the maximum likelihood method. 
Hence, in a nonparametrizable case, it depends on the case
which estimation makes the generalization loss smaller. 

\section{Conclusion}

In this paper, we theoretically proved that
equations of states in statistical estimation hold
even if a true distribution is nonparametrizable and regular 
for a learning machine. In the previous paper, we proved that
the equations of states hold even if a true distribution is
parametrizable and singular. By combining these results,
the equations of states are applicable without regard to
the condition of the true distribution and the learning machine.
Moreover, the equations of states contains AIC and TIC in the
special cases.

\subsection*{Acknowledgment}
This research was partially supported by the Ministry of Education,
 Science, Sports and Culture in Japan, Grant-in-Aid for Scientific
 Research 18079007. 

\section{Appendix}

\subsection{Proof of Lemma \ref{Lemma:111}}

By using eq.(\ref{eq:nabla-0-2}), $\nabla L_{n}(\hat{w})=0$, 
in a neighborhood of $\hat{w}$, 
\[
L_{n}(w)= L_{n}(\hat{w}) +
\frac{1}{2}(w-\hat{w})\cdot K_{n}(\hat{w})(w-\hat{w})
+r(w),
\]
where $r(w)$ is given by 
\[
r(w)=\frac{1}{6}\sum_{i,j,k=1}^{d} (\nabla^{3}L_{n}(\hat{w}))_{ijk}(w-\hat{w})_{i}(w-\hat{w})_{j}
(w-\hat{w})_{k} + O(|w-\hat{w}|^{4}).
\]
Hence, for a given function $g(w)$, 
the average by the {\it a posteriori} distribution is given by 
\begin{eqnarray*}
E_{w}[g(w)]
&=&\frac{
\int g(w)\;\exp\Bigl(
-\frac{n\beta}{2}(w-\hat{w})\cdot K_{n}(\hat{w})(w-\hat{w})
-n\beta r(w)
\Bigr)dw}{
\int \exp\Bigl(
-\frac{n\beta}{2}(w-\hat{w})\cdot K_{n}(\hat{w})(w-\hat{w}) 
-n\beta r(w)
\Bigr)dw}.
\end{eqnarray*}
The main region of the integration is a neighborhood of 
$\hat{w}$, $|w-\hat{w}|<\epsilon$, hence by putting $w'=\sqrt{n}(w-\hat{w})$, 
\[
E_{w}[g(w)]=\frac{
\int g(\hat{w}+\frac{w'}{\sqrt{n}})\;\exp(
-\frac{\beta}{2}w'\cdot K_{n}(\hat{w})w'-\frac{\beta\delta(w')}{\sqrt{n}}
+O_{p}(\frac{1}{n})
)dw'}{
\int \exp(
-\frac{\beta}{2}w'\cdot K_{n}(\hat{w})w'-\frac{\beta\delta(w')}{\sqrt{n}}
+O_{p}(\frac{1}{n})
)dw'},
\]
where $\delta(w')$ is the third-order polynomial, 
\[
\delta(w') =\frac{1}{6} \sum_{i,j,k=1}^{d}
(\nabla^{3}L_{n}(\hat{w}))_{ijk}w'_{i}w'_{j}w'_{k} .
\]
By using
\[
\exp\Bigl(-\frac{\beta\delta(w')}{\sqrt{n}}
+O_{p}(\frac{1}{n})\Bigr)=
1-\frac{\beta\delta(w')}{\sqrt{n}}
+O_{p}(\frac{1}{n}),
\]
it follows that 
\[
E_{w}[g(w)]=\frac{
\int g(\hat{w}+\frac{w'}{\sqrt{n}})\;
(1-\frac{\beta\delta(w')}{\sqrt{n}}
)
\exp(
-\frac{\beta}{2}w'\cdot K_{n}(\hat{w})w'
)dw'}{
\int 
\exp(
-\frac{\beta}{2}w'\cdot K_{n}(\hat{w})w'
)dw'}+O_{p}(\frac{1}{n}).
\]
Hence by putting $g(w)=w-\hat{w}$, we obtain eq.(\ref{eq:w-w0}) 
and by putting $g(w)=(w-\hat{w})(w-\hat{w})^{T}$, eq.(\ref{eq:w-w0-2}).
By the same way, 
eq.(\ref{eq:w-w0-3}) and
eq.(\ref{eq:w-w0-4}) are proved. 
Let us prove eq.(\ref{eq:e(w-0)(w-0)}). By using eq.(\ref{eq:w-w0-2}), 
\begin{eqnarray}
&& E_{w}[(w-w_{0})(w-w_{0})^{T}]\nonumber\\
&&=E_{w}[(\hat{w}-w_{0}+\frac{w'}{\sqrt{n}})(\hat{w}-w_{0}+\frac{w'}{\sqrt{n}})^{T}]\nonumber\\
&&=(\hat{w}-w_{0})(\hat{w}-w_{0})^{T}+\frac{1}{n}E_{w}[w'w'^{T}]+O_{p}(\frac{1}{n^{3/2}})\nonumber\\
&& =(\hat{w}-w_{0})
(\hat{w}-w_{0})^{T}+
\frac{K_{n}(\hat{w})^{-1}}{n\beta}+O_{p}(\frac{1}{n^{3/2}}). \label{eq:(w-w0)2}
\end{eqnarray}
Then by applying eq.(\ref{eq:JIJ}), eq.(\ref{eq:e(w-0)(w-0)}) is obtained.
Lastly, in general, 
\[
|w-w_{0}|^{3}\leq 3(|w-\hat{w}|^{3} + |\hat{w}-w_{0}|^{3}).
\]
Then, by eq.(\ref{eq:MAP-TRUE}) and eq.(\ref{eq:w-w0-4}), eq.(\ref{eq:e(w-0)3}) is derived. 
Therefore we have obtained Lemma 1. 

\subsection{Proof of Lemma \ref{Lemma:222} }

By the Taylor expansion of $f(X,w)$ at $w_{0}$, 
\begin{eqnarray}
f(X,w)
&=&\nabla f(X,w_{0}) \cdot(w-w_{0})\nonumber \\
&&+\frac{1}{2}(w-w_{0})\cdot \nabla^{2} f(X|w_{0})(w-w_{0}) \nonumber \\
&& + f_{3}(X,w),\label{eq:f(x,w)-T}
\end{eqnarray}
where $f_{3}(X,w)$ satisfies 
\[
|f_{3}(X,w)|\leq C(X,w)|w-w_{0}|^{3}
\]
in a neighborhood of $w_{0}$ with a function $C(X,w)\geq 0$. 
Let us estimate $D_{1},...,D_{6}$. Firstly, 
by using eq.(\ref{eq:nabla-0-1}) and eq.(\ref{eq:e(w-0)3}), 
\begin{eqnarray*}
D_{1}&=& \frac{1}{2}EE_{w}E_{X}[(w-w_{0})\cdot\nabla^{2} f(X,w_{0})(w-w_{0})]
+o(\frac{1}{n}) \\
&=&\frac{1}{2}EE_{w}[(w-w_{0})\cdot J(w-w_{0})]+o(\frac{1}{n}). 
\end{eqnarray*}
Then by using the identity
\[
(\forall u,v \in {\bf R}^{d}, A\in{\bf R}^{d\times d})\;\;\;u\cdot Av =\mbox{tr}(A vu^{T}),
\]
and eq.(\ref{eq:e(w-0)(w-0)}),
\begin{eqnarray*}
D_{1}
&=& \frac{1}{2}EE_{w}[\mbox{tr}((J(w-w_{0}))(w-w_{0})^{T})]+o(\frac{1}{n})\\
&=&
\frac{d}{2n\beta}+\frac{\mbox{tr}(IJ^{-1})}{2n}+o(\frac{1}{n}).
\end{eqnarray*}
Secondly, by using the identity
\[
(\forall u,v \in {\bf R}^{d})\;\;\;(u\cdot v)^{2} =\mbox{tr}((uu^{T})(vv^{T})),
\]
the definition of $I$, and eq.(\ref{eq:e(w-0)(w-0)}),
\begin{eqnarray*}
D_{2}&=& (1/2)E E_{w}[\;E_{X}[(\nabla f(X,w_{0})\cdot(w-w_{0}))^{2}]\;]+o(\frac{1}{n})\\
&=&(1/2)E E_{w}[\mbox{tr}(I(w-w_{0})(w-w_{0})^{T})]+o(\frac{1}{n})\\
&=& 
\frac{\mbox{tr}(IJ^{-1})}{2n\beta}+\frac{\mbox{tr}(IJ^{-1}IJ^{-1})}{2n} +o(\frac{1}{n}).
\end{eqnarray*}
Thirdly, by the definition of $I$, eq.(\ref{eq:w-w0}), and eq.(\ref{eq:JIJ}),
\begin{eqnarray*}
D_{3}&=& (1/2)E E_{X}[E_{w}[\nabla f(X,w_{0})\cdot(w-w_{0})]^{2}]+o(\frac{1}{n})\\
&=&(1/2)E E_{X}[(\nabla f(X,w_{0}) \cdot(\hat{w}-w_{0}))^{2}]+o(\frac{1}{n})\\
&=&(1/2)E [\mbox{tr}(I(\hat{w}-w_{0})(\hat{w}-w_{0})^{T})]+o(\frac{1}{n})\\
&=& 
\frac{\mbox{tr}(IJ^{-1}IJ^{-1})}{2n}+o(\frac{1}{n}).
\end{eqnarray*}
Fourthly, by the Taylor expansion eq.(\ref{eq:f(x,w)-T})
\begin{eqnarray*}
D_{4}&=&E E_{w}[ E_{j}^{(n)}[\nabla f(X_{j},w_{0})\cdot(w-w_{0})]]\\
&& +\frac{1}{2}
E E_{w}[ E_{j}^{(n)}[(w-w_{0})\cdot \nabla^{2} f(X_{j},w_{0})(w-w_{0})]]
+o(\frac{1}{n})\\
&=&E  [ \; E_{j}^{(n)}[\nabla f(X_{j},w_{0})]\cdot E_{w}[w-w_{0}]\;]\\
&& +\frac{1}{2}
E E_{w}[(w-w_{0})\cdot J_{n}(w_{0})(w-w_{0})]]+o(\frac{1}{n}).
\end{eqnarray*}
Then by using $E[J_{n}(w_{0})]=J$, the second term is equal to $D_{1}$. 
To the first term, we apply eq.(\ref{eq:w-w0}) and 
\[
E_{j}^{(n)}[\nabla f(X_{j},w_{0})]=\nabla L_{n}(w_{0})+O_{p}(\frac{1}{n}),
\]
we obtain 
\begin{eqnarray*} 
D_{4}
&=& E [ (\nabla L_{n}(w_{0}))\cdot(\hat{w}-w_{0})] 
+D_{1}+o(\frac{1}{n}).
\end{eqnarray*}
Then applying eq.(\ref{eq:nablaln}), $K_{n}(w_{0})\rightarrow J$, 
and $w_{\beta}^{*}\rightarrow w_{0}$,
\begin{eqnarray*}
D_{4}&=&
-E [(K_{n}(w_{\beta}^{*})(\hat{w}-w_{0}))\cdot (\hat{w}-w_{0})]
+D_{1}+o(\frac{1}{n})\\&=&
\frac{d}{2n\beta}-\frac{\mbox{tr}(IJ^{-1})}{2n}+o(\frac{1}{n}).
\end{eqnarray*}
And lastly, by the definitions, 
\begin{eqnarray*}
D_{5}&=& (1/2)E E_{w}[\;E_{j}^{n}[(\nabla f(X_{j},w_{0})\cdot(w-w_{0}))^{2}]\;]+o(\frac{1}{n}),\\
D_{6}&=& (1/2)E E_{j}^{n}[E_{w}[\nabla f(X_{j},w_{0})\cdot(w-w_{0})]^{2}]+o(\frac{1}{n}).
\end{eqnarray*}
By using the convergences in probability,
$I_{n}(w_{0})\rightarrow I$ and $J_{n}(w_{0})\rightarrow J$, 
it follows that 
\begin{eqnarray*}
D_{5}&=&D_{2}+o(\frac{1}{n}), \\
D_{6}&=&D_{3}+o(\frac{1}{n}),
\end{eqnarray*}
which completes Lemma 2.

\end{document}